\documentclass[letterpaper,10pt,conference]{ieeeconf}

\IEEEoverridecommandlockouts
\usepackage{graphicx}
\usepackage{cite}
\usepackage{url}
\usepackage{amsmath}
\usepackage{amssymb}
\usepackage{newtxmath}
\usepackage{booktabs}
\usepackage{array}
\usepackage{xspace}

\newcommand{\method}{CoWear\xspace}

\title{\LARGE \bf
CoWear: Towards Collaborative Wearable Inertial Localization with Multiple Smart Devices
}

\author{
Zhuoyuan She$^{1,*}$,
Zhanpeng Zhang$^{1,*}$,
Yapeng Chen$^{2}$,
Jiale Wang$^{3}$,
Xin Xie$^{4}$,
Ming Xia$^{3}$
\thanks{
$^{*}$Zhuoyuan She and Zhanpeng Zhang contributed equally to this work.
}
\thanks{
$^{1}$School of Electronic and Information Engineering,
Beihang University, Beijing, China.
}
\thanks{
$^{2}$School of Automation Science and Electrical Engineering,
Beihang University, Beijing, China.
}
\thanks{
$^{3}$School of Space and Earth Sciences,
Beihang University, Beijing, China.
}
\thanks{
$^{4}$Jiangxi Research Institute,
Beihang University, Nanchang, China.
}
}

\begin{document}
\maketitle
\thispagestyle{empty}
\pagestyle{empty}

\begin{abstract}
Consumer smart devices observe the same human motion from different body locations, yet existing inertial-localization benchmarks rarely provide synchronized heterogeneous wearables with device-specific motion ground truth. We present \method, a dataset and reference framework for collaborative inertial localization using a smartphone, smartwatch, and smart glasses. The dataset contains over 30 participants, over 300 walking sessions, and more than 30~hours and 100~km of aggregate device trajectories, with each device individually tracked by motion capture. To enable collaboration across distinct body motions, we train role-specific estimators to predict the displacement of a common reference device over synchronized temporal windows. Their predictions are transformed to the world frame using gyroscope integration and fused according to learned uncertainty. Experiments using phone, glasses, and watch as reference targets show that multi-device fusion consistently improves both displacement and trajectory rollout accuracy over individual-device estimation. \method provides a dataset and reproducible baseline for studying collaborative inertial localization across heterogeneous consumer wearables.
\end{abstract}

\section{Introduction}

Consumer devices such as smartphones, smartwatches, and smart glasses increasingly accompany users throughout daily activities and continuously measure motion with embedded inertial sensors. Their independence from external infrastructure makes them attractive for pedestrian localization in indoor and GNSS-denied environments, where learning-based inertial odometry has shown strong potential with low-cost IMUs \cite{chen2018ionet,herath2020ronin,liu2020tlio}. Yet most existing systems localize from one sensing device at a time, even though a user may simultaneously carry several inertial devices at different body locations. This naturally raises the question of whether these heterogeneous devices can provide redundant observations of human locomotion and jointly improve localization.

\begin{figure}[t]
    \centering
    \includegraphics[width=\linewidth]{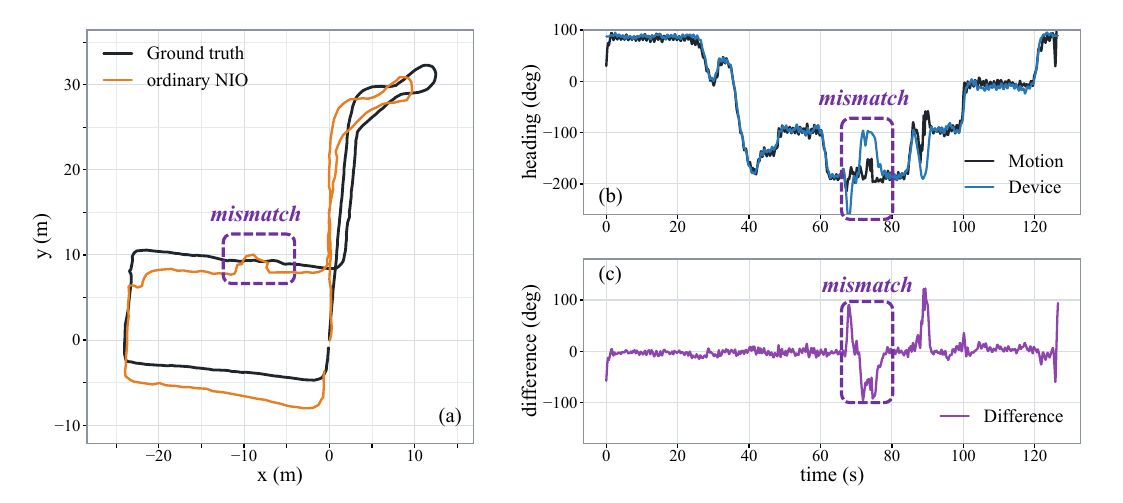}
    \caption{
    Representative device--motion mismatch in RIDI.
    (a) Ground-truth and NIO trajectories.
    (b) Motion and device headings.
    (c) Heading difference.
    The purple boxes highlight a short interval where device motion deviates sharply from the underlying locomotion.
    }
    \label{fig:device_body_mismatch}
\end{figure}

A key difficulty is that the motion of a sensing device is not necessarily the motion that localization aims to recover. A handheld phone, wrist-worn watch, and head-mounted glasses share the same global locomotion but also undergo distinct local motions caused by arm swing, head motion, and changing carrying styles. This discrepancy is already present in common inertial-localization settings where the IMU and trajectory reference correspond to different physical carriers \cite{yan2018ridi,herath2020ronin}. In such cases, device-specific motion is not directly supervised and must be implicitly absorbed by the estimator. As illustrated in Fig.~\ref{fig:device_body_mismatch}, a representative RIDI example highlights how substantial device motion can coexist with much smaller changes in the underlying human trajectory. 

This observation motivates two design choices in \method. First, multiple wearables provide redundant but non-identical observations of the same locomotion, offering the possibility of reducing the influence of device-specific disturbances. Second, to study this redundancy explicitly, the motion of each sensing device itself must be observable. We therefore collect synchronized smartphone, smartwatch, and smart-glasses inertial data while independently tracking every device with motion capture. This produces device-specific 6-DoF ground truth and allows each wearable to be analyzed either as a motion source or as a common localization reference.

The remaining challenge is how to compare and combine observations from non-co-located body locations. For every synchronized temporal window, role-specific estimators predict the displacement of one selected reference device from each wearable. These estimates are transformed into a shared world frame using gyroscope integration and combined through learned uncertainty. This common-target formulation avoids averaging trajectories of different physical carriers. Experiments with phone-, watch-, and glasses-reference targets show consistent improvements from collaborative estimation over single-device inference.

\begin{figure*}[t]
    \centering
    \includegraphics[width=\textwidth]{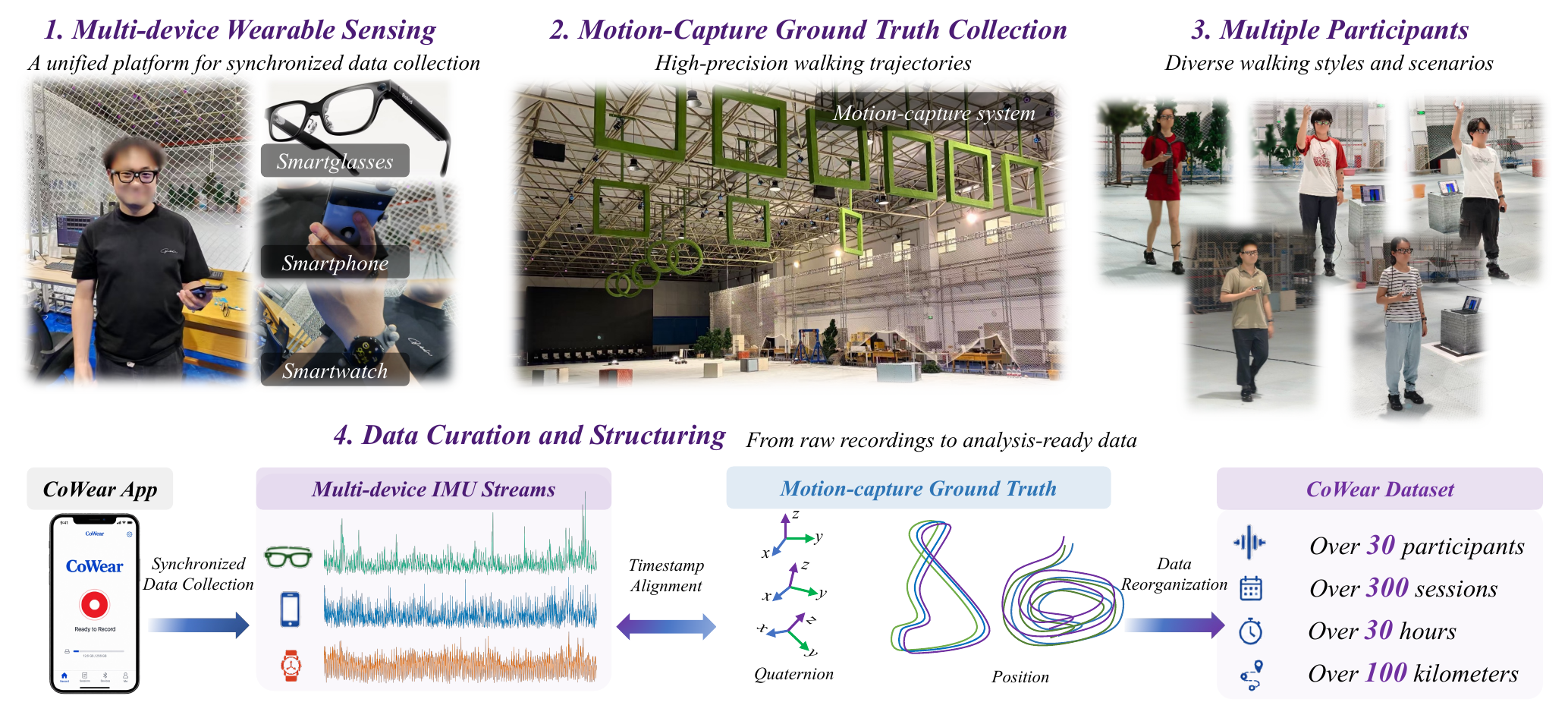}
    \caption{
    Overview of the \method dataset construction pipeline.
    Three heterogeneous wearables are recorded simultaneously and independently
    tracked by motion capture. Data from multiple participants are temporally
    aligned, reorganized, and curated into synchronized analysis-ready sessions.
    Dataset scale is reported using aggregate device trajectories.
    }
    \label{fig:overview}
\end{figure*}

Our contributions are as follows:
\begin{itemize}
\item We introduce \method, a synchronized multi-device inertial dataset containing smartphone, smartwatch, and smart-glasses measurements with device-specific 6-DoF motion-capture ground truth.

\item We formulate collaborative localization across non-co-located wearables as common-reference displacement estimation from synchronized heterogeneous observations.
\item We provide a reproducible collaborative localization baseline that maps heterogeneous wearable observations to a common reference displacement and combines them through uncertainty-aware fusion.

\end{itemize}

\section{Related Work}
\label{sec:related}

\paragraph{Neural inertial localization.}
Classical pedestrian dead reckoning constrains inertial drift using footfall timing, heading estimation, and stride models \cite{foxlin2005pedestrian,ojeda2007nongps,weinberg2002adxl202}.
IONet introduced short-window neural inertial odometry by directly regressing motion from inertial sequences, while RoNIN improved robustness to varying phone placements through learned velocity estimation \cite{chen2018ionet,herath2020ronin}.
TLIO further predicts displacement together with uncertainty and integrates learned estimates into a filtering framework, whereas NILoc extends learned inertial reasoning toward scene-level localization \cite{liu2020tlio,herath2022niloc}.
More recent methods explore stronger temporal modeling and geometric structure, including HNNTA, IMUNet, and equivariant inertial odometry \cite{wang2022hnnta,zeinali2024imunet,jayanth2025eqnio}.
These approaches substantially advance single-device inertial localization, but generally formulate each sensing stream as an independent motion-estimation problem rather than exploiting simultaneous observations from heterogeneous wearables.

\paragraph{Datasets and supervision for inertial localization.}
RIDI provides smartphone inertial measurements together with reference trajectories for learning pedestrian motion, while OxIOD further covers diverse carrying placements and motion modes for deep inertial odometry \cite{yan2018ridi,chen2018oxiod}.
RoNIN provides large-scale free-form smartphone inertial data with body-referenced motion supervision, and TLIO supports learning displacement and uncertainty for inertial odometry \cite{herath2020ronin,liu2020tlio}.
ADVIO complements these datasets with challenging visual--inertial pedestrian navigation sequences \cite{cortes2018advio}.
Although these datasets adopt different sensing and supervision configurations, in several common settings the IMU measurements and trajectory reference correspond to different physical carriers, leaving device-to-reference motion implicit.
In contrast, \method independently tracks every consumer device, pairing each IMU stream with the 6-DoF motion of the same physical carrier and enabling explicit analysis of device-specific motion.

\paragraph{Multi-wearable and body-distributed sensing.}
Suite-IN studies inertial navigation using multiple Apple consumer devices, while Project Aria provides a head-worn multimodal sensing platform with globally aligned measurements and Aria Everyday Activities extends this setting to daily human activities \cite{sun2024suitein,engel2023projectaria,lv2024ariaeveryday}.
DIP reconstructs human pose from sparse body-mounted IMUs, and TransPose further estimates both pose and global translation from distributed inertial sensors \cite{huang2018dip,yi2021transpose}.
HelmetPoser investigates head-mounted inertial motion with corresponding motion ground truth \cite{li2025helmetposer}.
These works demonstrate the value of body-distributed sensing, but their primary goals differ from collaborative inertial localization across heterogeneous consumer devices.
\method instead focuses on preserving the distinct motion of each wearable while allowing multiple devices to jointly estimate a common localization target.

\section{CoWear Dataset}
\label{sec:dataset}

The construction of \method is summarized in Fig.~\ref{fig:overview}.
The dataset is built around four stages: synchronized sensing from heterogeneous
consumer wearables, device-specific motion-capture tracking, data collection
across multiple participants and walking conditions, and fine-grained temporal
alignment and curation. A central design principle is that each IMU stream is
paired with the motion of the same physical device, allowing device-specific
motion to be observed rather than implicitly absorbed into a body-level
reference trajectory.

\subsection{Data Collection}
\label{subsec:data_collection}

\method uses three representative consumer devices: Rokid smart glasses, a
Google Pixel~6 smartphone, and a Google Pixel Watch~2 smartwatch. Accelerometer
and gyroscope measurements from all three devices are recorded at 100~Hz, while
a motion-capture (MoCap) system independently tracks the 6-DoF pose of each
device at 120~Hz. We developed the \emph{CoWear App} to issue unified recording
commands across the three devices, providing coarse synchronization during data
collection.

Unlike datasets in which inertial measurements and trajectory supervision may
refer to different physical carriers, each device in \method is equipped with
its own MoCap tracking configuration. The smartphone, smartwatch, and smart
glasses therefore each have an independent position and orientation trajectory
corresponding to the same device from which the IMU measurements are obtained.
This device-specific supervision preserves the distinct motion characteristics
of different body locations and enables each wearable to be analyzed either as
an independent motion source or as a reference target for collaborative
localization.

The collection protocol focuses on natural walking behavior. Participants are
instructed to face approximately along the direction of travel, hold the
smartphone approximately level, and wear the smart glasses normally. We include
two phone--watch configurations. In the \emph{same-hand} configuration, the
smartphone is held by the same arm on which the watch is worn, causing the two
devices to share more local hand motion. In the \emph{different-hand}
configuration, the smartphone is held in one hand while the watch is worn on the
opposite wrist, allowing the watch to undergo natural arm swing relative to the
phone.

The dataset contains over 30 participants and over 300 walking sessions.
Counting the three device trajectories separately, \method contains more than
30~hours and 100~km of aggregate device trajectories. These values describe the
total amount of device-level motion data rather than the physical walking
duration or distance of the participants.

Table~\ref{tab:dataset_comparison} compares \method with representative
inertial-motion datasets in terms of sensing setup, scale, supervision, and
joint-device support. Unlike prior resources, \method combines device-specific
motion supervision with synchronized heterogeneous wearables, enabling explicit
analysis of device motion and collaborative localization.

\begin{table*}[t]
\centering
\caption{Comparison with representative inertial-motion datasets.}
\label{tab:dataset_comparison}
\scriptsize
\setlength{\tabcolsep}{3.8pt}
\renewcommand{\arraystretch}{1.12}

\begin{tabular}{
l
p{0.15\textwidth}
c
c
p{0.16\textwidth}
p{0.16\textwidth}
c
}
\toprule
\textbf{Dataset}
&
\textbf{IMU setup}
&
\textbf{Duration}
&
\textbf{Distance}
&
\textbf{Reference source}
&
\textbf{Supervision target}
&
\textbf{Joint streams}
\\
\midrule

RIDI~\cite{yan2018ridi}
&
1 phone, 4 placements
&
2.5 h
&
--
&
Tango phone
&
Body trajectory
&
$\times$
\\

OxIOD~\cite{chen2018oxiod,chen2020lionet}
&
1 phone, 4 placements
&
14.7 h
&
42.5 km
&
MoCap / Tango
&
Phone / body
&
$\times$
\\

RoNIN~\cite{herath2020ronin}
&
1 free-carried phone
&
42.7 h
&
--
&
Tracking phone
&
Body trajectory
&
$\times$
\\

TLIO~\cite{liu2020tlio}
&
Headset IMU--camera
&
--
&
--
&
Headset VIO
&
Headset 6-DoF
&
$\times$
\\

DIP-IMU~\cite{huang2018dip}
&
17 body IMUs
&
--
&
--
&
Optical MoCap
&
Full-body pose
&
$\checkmark$ Homogeneous
\\

HelmetPoser~\cite{li2025helmetposer}
&
Helmet-mounted IMUs
&
--
&
--
&
VICON MoCap
&
Head 6-DoF
&
$\checkmark$ Homogeneous
\\

\midrule

\method
&
Phone + watch + glasses
&
$>$30 h
&
$>$100 km
&
\textbf{Device-specific MoCap}
&
\textbf{Device 6-DoF}
&
\textbf{$\checkmark$ Heterogeneous}
\\

\bottomrule
\end{tabular}
\end{table*}

\subsection{Data Curation and Structuring}
\label{subsec:data_curation}

The unified recording commands issued by the CoWear App provide only coarse
synchronization. The three consumer devices maintain independent clocks, and
communication and command-execution delays introduce device-dependent temporal
offsets. The MoCap system also operates on an independent clock. We therefore
perform fine temporal alignment by matching rotational motion observed by the
onboard gyroscope with that derived from the corresponding MoCap orientation
trajectory.

For device $r$, let $\mathbf{R}^{r}(t)\in\mathrm{SO}(3)$ denote its orientation
measured by MoCap. A pseudo angular velocity is obtained from consecutive
orientation samples as
\begin{equation}
    \tilde{\boldsymbol{\omega}}^{r}(t)
    =
    \frac{1}{\Delta t}
    \operatorname{Log}
    \left(
        \mathbf{R}^{r}(t)^{\top}
        \mathbf{R}^{r}(t+\Delta t)
    \right)^{\vee},
    \label{eq:pseudo_angular_velocity}
\end{equation}
where $\operatorname{Log}(\cdot)$ maps a rotation matrix to its Lie algebra and
$(\cdot)^{\vee}$ converts the resulting skew-symmetric matrix into a
three-dimensional rotation vector.

The temporal offset is then estimated by maximizing the correlation between the
magnitudes of the measured and MoCap-derived angular velocities,
\begin{equation}
    \tau_r^{*}
    =
    \arg\max_{\tau}
    \operatorname{Corr}
    \left(
        \left\|\boldsymbol{\omega}^{r}(t)\right\|_2,
        \left\|\tilde{\boldsymbol{\omega}}^{r}(t+\tau)\right\|_2
    \right),
    \label{eq:temporal_alignment}
\end{equation}
where $\boldsymbol{\omega}^{r}(t)$ is the angular velocity measured by the
onboard IMU. Using angular-velocity magnitude makes the alignment insensitive to
the coordinate-frame difference between the onboard IMU and the MoCap system.

The smartphone is first used to align the CoWear recording timeline with the
MoCap timeline. The same procedure is subsequently applied to the smart glasses
and smartwatch to compensate for residual device-specific offsets. This places
all IMU measurements and device trajectories on a common temporal reference.

After alignment, all streams are resampled to 100~Hz. Recordings containing
MoCap tracking gaps longer than 2~s are discarded. Short missing position
segments are recovered by linear interpolation, while orientation gaps are
filled using spherical linear interpolation (SLERP). The resulting synchronized
streams are reorganized into session-level records while preserving the
individual trajectories of the smartphone, smartwatch, and smart glasses.

\section{Benchmark Tasks and Collaborative Localization Framework}
\label{sec:benchmark}

\subsection{Benchmark Task Formulation}

Let $\mathcal{R}=\{\mathrm{phone},\mathrm{watch},\mathrm{glasses}\}$ denote the
set of wearable roles. We use $r\in\mathcal{R}$ to denote an observing device,
$q\in\mathcal{R}$ to denote the physical target to be localized, and $k$ to
index temporal windows. For the $k$-th temporal window
$\mathcal{W}_k=[t_k,t_{k+1}]$, the inertial observation from device $r$ is
written as $X_k^r=\{\mathbf{x}_t^r\}_{t\in\mathcal{W}_k}$, where
$\mathbf{x}_t^r=[\boldsymbol{\omega}_t^r,\mathbf{a}_t^r]\in\mathbb{R}^{6}$.
Here, $\boldsymbol{\omega}_t^r\in\mathbb{R}^{3}$ and
$\mathbf{a}_t^r\in\mathbb{R}^{3}$ denote the raw gyroscope and accelerometer
measurements, respectively. All inertial measurements remain in their native
device frames.

CoWear defines two complementary localization tasks.

\paragraph{Task 1: Single-device localization.}
Given the inertial observation $X_k^r$ from device $r$, an estimator predicts
the displacement of that same physical device over the window:
\begin{equation}
\Delta\mathbf{p}_{k,r}^{W}
=
\mathbf{p}_{r}^{W}(t_{k+1})
-
\mathbf{p}_{r}^{W}(t_k),
\label{eq:single_device_task}
\end{equation}
where $\mathbf{p}_{r}^{W}(t)\in\mathbb{R}^{3}$ denotes the motion-capture
position of device $r$ in the world frame $W$. Since each wearable is paired
with its own motion-capture trajectory, this task provides device-aligned
supervision and allows existing inertial-localization methods to be evaluated
consistently across phone-, wrist-, and head-worn sensing.

\paragraph{Task 2: Collaborative localization.}
The collaborative task asks whether synchronized observations from multiple
wearables can jointly estimate the motion of one selected physical target
$q$. The ground-truth displacement of target $q$ over the same temporal window
is
\begin{equation}
\Delta\mathbf{p}_{k,q}^{W}
=
\mathbf{p}_{q}^{W}(t_{k+1})
-
\mathbf{p}_{q}^{W}(t_k).
\label{eq:target_displacement}
\end{equation}
Each observing device $r$ independently estimates this same
$\Delta\mathbf{p}_{k,q}^{W}$ from its own inertial observation $X_k^r$.

To enable direct fusion of displacement estimates, we adopt a
common-target formulation that accounts for the non-co-located
nature of wearable devices. Although a phone, watch, and pair
of glasses observe the same underlying human locomotion,
they follow different physical trajectories due to their
body locations and local motions. Directly averaging these
device-specific trajectories is therefore not physically
well-defined. Instead, heterogeneous inertial observations
are mapped to the motion of the same physical target,
making their predictions directly comparable and suitable
for fusion.

For prediction in the native coordinate frame of observing device $r$, the
target displacement is represented as
\begin{equation}
\Delta\mathbf{p}_{k,q}^{I_r}
=
\left(
\mathbf{R}_{I_r}^{W}(t_k)
\right)^{\top}
\Delta\mathbf{p}_{k,q}^{W},
\label{eq:local_target}
\end{equation}
where $I_r$ denotes the local coordinate frame of device $r$ and
$\mathbf{R}_{I_r}^{W}(t_k)\in SO(3)$ denotes the rotation from $I_r$ to the
world frame $W$ at the beginning of the window. In this formulation, all
observing devices are supervised by the same physical target while retaining
their own sensing coordinates and device-specific motion characteristics.

\subsection{Collaborative Localization Framework}

\begin{figure}[t]
    \centering
    \includegraphics[width=\columnwidth]{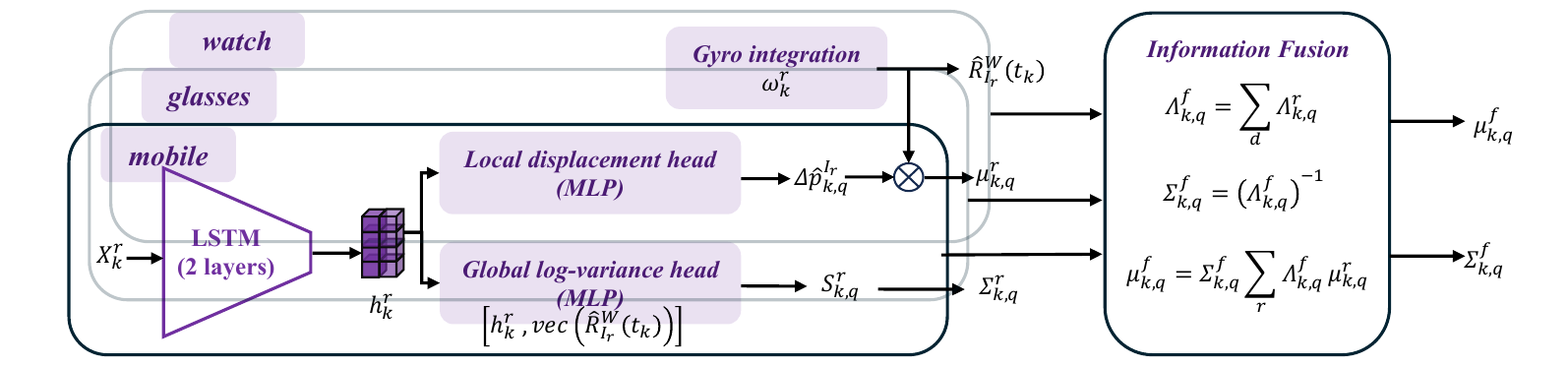}
    \caption{Collaborative localization framework. Each wearable independently
    estimates the same target displacement and its uncertainty before fusion.}
    \label{fig:localization_framework}
\end{figure}

To instantiate the collaborative localization task, we provide the framework
shown in Fig.~\ref{fig:localization_framework}. Each wearable is processed
independently by the same estimation pipeline. Its native-frame inertial
sequence is encoded into a motion representation, from which a displacement
head predicts the motion of the common target in the observing device frame.
The prediction is subsequently transformed into the world frame using the
orientation obtained from gyroscope integration. In parallel, an uncertainty
head estimates the reliability of the corresponding world-frame prediction for
multi-device fusion.

For each observing device $r$, the 6-D inertial sequence $X_k^r$ is first
projected into a 128-D feature space and then processed by a two-layer
unidirectional LSTM. The resulting motion representation is passed to a
displacement head, which predicts
$\widehat{\Delta\mathbf{p}}_{k,q}^{I_r}\in\mathbb{R}^{3}$, namely the
displacement of target $q$ expressed in the local frame $I_r$ of the observing
device.

The orientation of each observing device is propagated from its gyroscope
measurements. Starting from an initial orientation provided at the beginning of
each rollout, the orientation is updated as
\begin{equation}
\mathbf{R}_{I_r}^{W}(t+\Delta t)
=
\mathbf{R}_{I_r}^{W}(t)
\operatorname{Exp}
\left(
\boldsymbol{\omega}_t^r\Delta t
\right),
\label{eq:gyro_propagation}
\end{equation}
where $\Delta t$ denotes the IMU sampling interval and
$\operatorname{Exp}(\cdot)$ denotes the exponential map from the rotation
vector to $SO(3)$. The predicted local displacement is then transformed into
the world frame:
\begin{equation}
\boldsymbol{\mu}_{k,q}^{r}
=
\mathbf{R}_{I_r}^{W}(t_k)
\widehat{\Delta\mathbf{p}}_{k,q}^{I_r},
\label{eq:world_prediction}
\end{equation}
where $\boldsymbol{\mu}_{k,q}^{r}\in\mathbb{R}^{3}$ denotes the world-frame
mean displacement predicted by observing device $r$ for target $q$.

In parallel, an uncertainty head predicts a three-dimensional log-variance
vector
$\mathbf{s}_{k,q}^{r}
=
[s_{k,q,x}^{r},s_{k,q,y}^{r},s_{k,q,z}^{r}]^{\top}$.
The corresponding world-frame covariance is represented as
$\boldsymbol{\Sigma}_{k,q}^{r}
=
\operatorname{diag}
(
e^{s_{k,q,x}^{r}},
e^{s_{k,q,y}^{r}},
e^{s_{k,q,z}^{r}}
)$.
The uncertainty head is conditioned on both the learned motion representation
and the current device orientation, allowing the estimated reliability to vary
across world-frame directions.

Each device estimator is trained with local displacement supervision together
with a heteroscedastic Gaussian objective in the world frame. Let
$\mathbf{e}_{k,q}^{r}
=
\boldsymbol{\mu}_{k,q}^{r}
-
\Delta\mathbf{p}_{k,q}^{W}$
denote the world-frame displacement error. The training objective for observing
device $r$ and target $q$ is
\begin{equation}
\begin{aligned}
\mathcal{L}_{r,q}
={}&
\operatorname{SmoothL1}
\left(
\widehat{\Delta\mathbf{p}}_{k,q}^{I_r},
\Delta\mathbf{p}_{k,q}^{I_r}
\right)
\\
&+
\frac{1}{2}
\sum_{j\in\{x,y,z\}}
\left[
(e_{k,q,j}^{r})^{2}
e^{-s_{k,q,j}^{r}}
+
s_{k,q,j}^{r}
\right].
\end{aligned}
\label{eq:training_loss}
\end{equation}
The uncertainty used during collaboration is therefore learned directly from
localization residuals rather than introduced as a manually specified fusion
weight.

\subsection{Uncertainty-Aware Multi-Device Fusion}

After common-target estimation, all available observing devices predict the
same physical displacement over the same temporal window and in the same world
frame. The prediction from observing device $r$ is modeled as
$\mathcal{N}
(
\boldsymbol{\mu}_{k,q}^{r},
\boldsymbol{\Sigma}_{k,q}^{r}
)$.

Let $\mathcal{V}_k\subseteq\mathcal{R}$ denote the set of observing devices
available during window $k$. We define the information matrix associated with
each prediction as
\begin{equation}
\boldsymbol{\Lambda}_{k,q}^{r}
=
\left(
\boldsymbol{\Sigma}_{k,q}^{r}
\right)^{-1}.
\label{eq:information_matrix}
\end{equation}
Assuming conditionally independent Gaussian estimates, their predictions are
fused in information form:

\begin{equation}
\begin{aligned}
\boldsymbol{\Sigma}_{k,q}^{\mathrm f}
&=
\left[
\sum_{r\in\mathcal{V}_k}
\left(
\boldsymbol{\Sigma}_{k,q}^{r}
\right)^{-1}
\right]^{-1},
\\
\boldsymbol{\mu}_{k,q}^{\mathrm f}
&=
\boldsymbol{\Sigma}_{k,q}^{\mathrm f}
\sum_{r\in\mathcal{V}_k}
\left(
\boldsymbol{\Sigma}_{k,q}^{r}
\right)^{-1}
\boldsymbol{\mu}_{k,q}^{r},
\end{aligned}
\label{eq:information_fusion}
\end{equation}
where the superscript $\mathrm f$ denotes the fused estimate. Since
$\boldsymbol{\Sigma}_{k,q}^{r}$ is diagonal, different observing devices can
contribute differently along the three world axes. If a wearable is
unavailable, it is simply excluded from $\mathcal{V}_k$.

The fused displacement is finally accumulated to recover the target trajectory:
\begin{equation}
\widehat{\mathbf{p}}_{q}^{W}(t_{k+1})
=
\widehat{\mathbf{p}}_{q}^{W}(t_k)
+
\boldsymbol{\mu}_{k,q}^{\mathrm f}.
\label{eq:trajectory_rollout}
\end{equation}
Each rollout uses only the initial position of target $q$ and the initial
orientation of each observing device. No intermediate ground-truth position or
orientation is used for correction.

\section{Experiments}
\label{sec:experiments}

\subsection{Experimental Setup}

\paragraph{Data split and training.}
We use all 363 synchronized sessions containing complete phone, watch, and
glasses recordings. The dataset is divided at the session level into 217
training sessions, 31 validation sessions, and 115 test sessions, and windows
from the same session never cross different splits.

The rigid transformation between each IMU and its corresponding
motion-capture marker is obtained through offline extrinsic calibration. Each
trajectory rollout is initialized using the calibrated initial orientation of
the observing device, after which orientation is propagated only through
gyroscope integration. No intermediate ground-truth position or orientation is
used to correct the rollout.

Motion sequences are partitioned into variable-length joint motion events
detected from the synchronized phone, watch, and glasses streams. Specifically,
event boundaries are obtained by peak--trough detection on the combined
acceleration magnitude of the three devices. The same event boundaries are
shared across all wearable observations so that they correspond to the same
temporal motion interval.

For the single-device benchmark, we evaluate a classical Weinberg
peak--trough PDR pipeline together with RIDI~\cite{yan2018ridi},
RoNIN~\cite{herath2020ronin}, TLIO~\cite{liu2020tlio}, and CoWear LSTM.
All learning-based baselines are retrained on CoWear using the same
train--validation--test split, while preserving their original model
formulations as closely as possible. CoWear LSTM is used as a lightweight
reference baseline rather than as a new single-device state-of-the-art method.

For benchmark late fusion, each method first performs self-localization
independently on the phone, watch, and glasses. The resulting self-trajectories
are aligned to a common origin and uniformly combined on a shared temporal
grid, with the fused trajectory evaluated against the phone target. CoWear LSTM
uses its three self-localization checkpoints under the same late-fusion
protocol as the external baselines.

For the common-target experiments, CoWear LSTM is trained so that observations
from different wearable roles predict the displacement of the same selected
physical target before fusion. The models are trained with AdamW, a batch size
of 1024, and at most 100 epochs. Hyperparameters and checkpoints are selected
using only the training and validation sets.

\paragraph{Metrics.}
We evaluate localization in the horizontal world $x$--$z$ plane. Let
$\Pi_{xz}(\cdot)$ denote projection onto the horizontal coordinates.

For each test session $s$, absolute trajectory error (ATE) is computed as
\begin{equation}
\mathrm{ATE}_{s}
=
\sqrt{
\frac{1}{N_s}
\sum_{k=1}^{N_s}
\left\|
\Pi_{xz}
\left(
\widehat{\mathbf{p}}_{q,s}^{W}(t_k)
-
\mathbf{p}_{q,s}^{W}(t_k)
\right)
\right\|_2^2
},
\label{eq:session_ate}
\end{equation}
where $N_s$ denotes the number of trajectory states in session $s$.

We report both the median and mean ATE across all $S=115$ test sessions:
\begin{equation}
\mathrm{ATE}_{\mathrm{med}}
=
\operatorname{median}_{s}\mathrm{ATE}_{s},
\qquad
\mathrm{ATE}_{\mathrm{mean}}
=
\frac{1}{S}
\sum_{s=1}^{S}
\mathrm{ATE}_{s}.
\label{eq:ate_statistics}
\end{equation}
This per-session evaluation gives each trajectory equal weight regardless of
its duration. For the common-target analysis, we additionally report
horizontal displacement RMSE at the event level. All errors are reported in
meters.

\begin{figure*}[t]
    \centering
    \includegraphics[width=\textwidth]{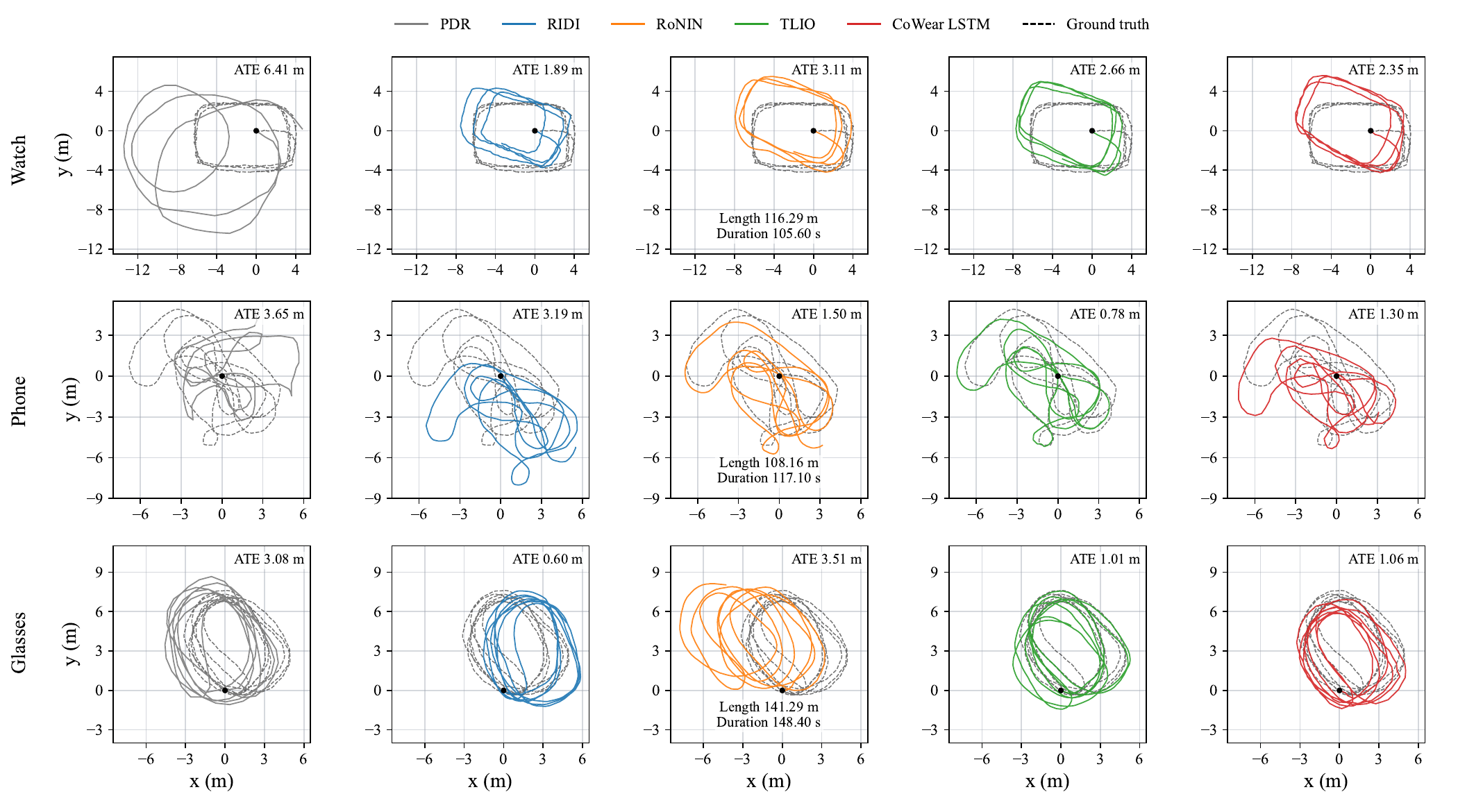}
    \caption{Representative benchmark trajectories for watch, phone, and
    glasses sequences. Dashed lines denote ground truth and solid lines denote
    predictions; per-session ATE is reported in each panel.}
    \label{fig:qualitative_trajectories}
\end{figure*}

\subsection{Benchmark Results}

Table~\ref{tab:benchmark_results} reports single-device localization and
uniform late-fusion performance on the 115 test sessions.

\begin{table}[t]
\centering
\caption{Trajectory ATE on the test sessions
(m, median / mean; lower is better).}
\label{tab:benchmark_results}
\footnotesize
\setlength{\tabcolsep}{3.2pt}
\begin{tabular}{@{}lcccc@{}}
\toprule
& \multicolumn{3}{c}{Single Device}
& \multicolumn{1}{c}{Late Fusion} \\
\cmidrule(lr){2-4}\cmidrule(l){5-5}
Method & Phone & Glasses & Watch & Fusion \\
\midrule
PDR
& 3.484 / 3.949
& 2.991 / 3.295
& 7.011 / 9.486
& 3.653 / 4.211 \\

RIDI~\cite{yan2018ridi}
& 1.570 / 1.825
& 1.710 / 2.016
& 1.840 / 2.236
& 1.329 / 1.532 \\

RoNIN~\cite{herath2020ronin}
& 1.720 / 2.361
& 2.997 / 3.234
& 2.137 / 2.668
& 1.837 / 2.096 \\

TLIO~\cite{liu2020tlio}
& \textbf{1.256 / 1.516}
& \textbf{1.543 / 1.647}
& 1.553 / 1.968
& \textbf{1.168 / 1.340} \\

CoWear LSTM
& 1.389 / 1.686
& 1.868 / 1.983
& 1.689 / \textbf{1.911}
& 1.313 / 1.441 \\
\bottomrule
\end{tabular}
\end{table}

Learning-based methods substantially outperform the classical peak--trough PDR
baseline across the three wearable roles. More importantly, the same method
can exhibit noticeably different behavior on the phone, watch, and glasses,
showing that heterogeneous wearable placements constitute distinct inertial
localization conditions rather than interchangeable sensor observations.

The late-fusion results further show that additional wearable observations do
not automatically improve localization. RIDI, TLIO, and CoWear LSTM benefit
from uniform late fusion, whereas PDR and RoNIN do not outperform their
strongest individual-device estimates. The same overall tendency is observed
under both median and mean ATE. This suggests that directly averaging
independently estimated self-trajectories is not a uniformly reliable strategy
for collaborative localization.

The purpose of this benchmark is therefore not to establish CoWear LSTM as a
new state-of-the-art inertial odometry model. Instead, CoWear LSTM serves as a
compact reference baseline, while the benchmark exposes how existing and
reference methods behave across heterogeneous wearable roles and under simple
multi-device fusion.

Figure~\ref{fig:qualitative_trajectories} provides qualitative examples of the
device-dependent behavior observed in Table~\ref{tab:benchmark_results}.
Different methods exhibit different amounts of drift, directional deviation,
and trajectory-shape distortion across the three wearable placements. These
examples complement the aggregate statistics and further illustrate why
multi-device inertial localization should explicitly account for differences
between wearable roles.

\begin{figure*}[t]
    \centering
    \includegraphics[width=\textwidth]{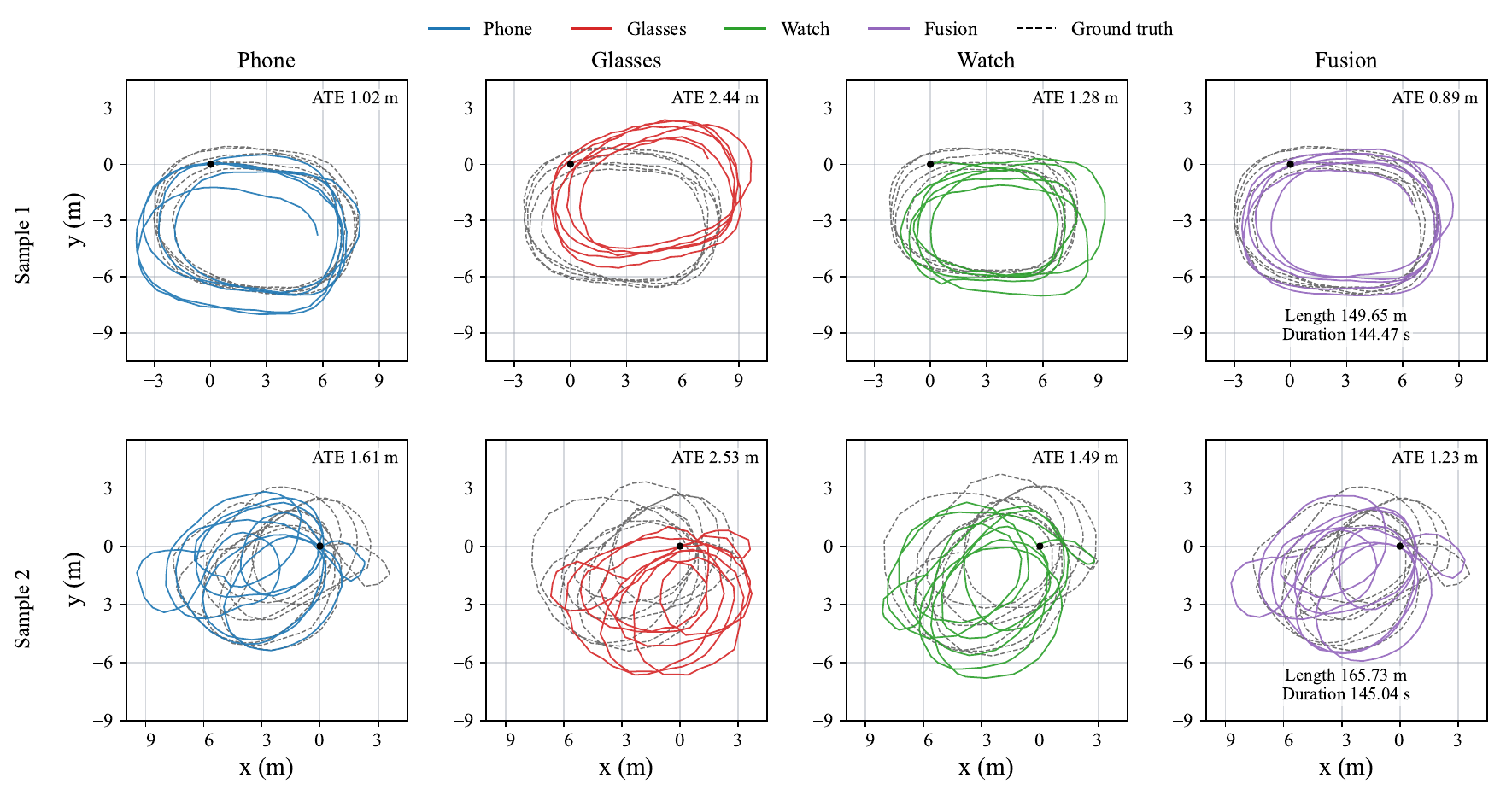}
    \caption{Representative common-target trajectories from individual
    wearable observations and information fusion. Fusion yields lower
    per-session ATE than each individual estimate in both examples.}
    \label{fig:fusion_trajectories}
\end{figure*}

\subsection{Collaborative Localization Analysis}

We next use CoWear LSTM to examine whether synchronized observations from
different body locations can estimate the motion of a common physical target
and whether these estimates can be combined effectively.

\paragraph{Common-target estimation.}
Table~\ref{tab:common_target} evaluates phone, glasses, and watch observations
against each of the three target devices. Within each target group, all
observers are trained to predict the displacement of the same physical target.
The fused result is obtained using the uncertainty-aware
information fusion described above.

\begin{table}[t]
\centering
\caption{Common-target localization across wearable roles
(m, lower is better).}
\label{tab:common_target}
\footnotesize
\setlength{\tabcolsep}{3.8pt}
\begin{tabular}{@{}llccc@{}}
\toprule
Target & Observation & Disp. RMSE & ATE Med. & ATE Mean \\
\midrule
Phone
& Phone
& 0.196 & 1.440 & 1.708 \\
& Glasses
& 0.209 & 1.934 & 2.050 \\
& Watch
& 0.210 & 1.671 & 2.004 \\
& \textit{Information fusion}
& \textbf{0.131} & \textbf{1.173} & \textbf{1.335} \\
\midrule

Glasses
& Phone
& 0.186 & 1.408 & 1.662 \\
& Glasses
& 0.195 & 1.868 & 1.983 \\
& Watch
& 0.208 & 1.738 & 1.995 \\
& \textit{Information fusion}
& \textbf{0.125} & \textbf{1.162} & \textbf{1.326} \\
\midrule

Watch
& Phone
& 0.220 & 1.543 & 1.829 \\
& Glasses
& 0.226 & 1.813 & 2.068 \\
& Watch
& 0.218 & 1.761 & 1.982 \\
& \textit{Information fusion}
& \textbf{0.149} & \textbf{1.164} & \textbf{1.368} \\
\bottomrule
\end{tabular}
\end{table}

The results show that the observing device does not need to coincide with the
target device. Phone inertial observations, for example, can estimate the
motion of both glasses and watch targets, while observations from the other
wearables can similarly recover phone motion. In several cases, an observation
from another body location even achieves lower trajectory error than the target
device's own observation.

This behavior demonstrates that synchronized wearable signals contain
complementary information about the same underlying locomotion process.
Device-aligned ground truth is important here because it allows observations
from different body locations to be supervised against the trajectory of a
common physical target.

Information fusion consistently improves over every individual observation
branch for all three targets. For the phone target, median ATE decreases from
1.440~m for the strongest individual branch to 1.173~m after fusion. The
corresponding improvements are 1.408~m to 1.162~m for the glasses target and
1.543~m to 1.164~m for the watch target. Mean ATE exhibits the same trend,
showing that the collaborative gain is consistent under both aggregation
statistics.

Figure~\ref{fig:fusion_trajectories} qualitatively illustrates this
collaborative effect. The strongest individual observer varies across
sequences, whereas the fused estimate remains closer to the target trajectory.
Together with Table~\ref{tab:common_target}, these results suggest that
observations from spatially distinct wearables contain complementary
information that can be exploited to improve localization accuracy.

\paragraph{Fusion strategy and observation diversity.}
Table~\ref{tab:collaboration_analysis} compares three fusion strategies.
Uniform fusion assigns the same contribution to all available observers.
Fixed-weight fusion uses device-specific reliability weights determined on the
validation set and keeps them unchanged during testing. Information fusion
instead uses the uncertainty predicted for each individual motion event and is
the default fusion strategy in CoWear.

\begin{table}[t]
\centering
\caption{Collaborative localization analysis
(ATE in m, median / mean; lower is better).}
\label{tab:collaboration_analysis}
\footnotesize
\setlength{\tabcolsep}{4.0pt}
\begin{tabular}{@{}lccc@{}}
\toprule
& Phone$^*$ & Glasses$^*$ & Watch$^*$ \\
\midrule
\multicolumn{4}{@{}l}{\textit{Fusion strategy}} \\

Uniform
& 1.359 / 1.429
& 1.294 / 1.390
& 1.304 / 1.433 \\

Fixed weight
& 1.308 / 1.411
& 1.195 / 1.376
& 1.252 / 1.424 \\

Information
& \textbf{1.173 / 1.335}
& \textbf{1.162 / 1.326}
& \textbf{1.164 / 1.368} \\

\midrule
\multicolumn{4}{@{}l}{\textit{Watch--phone relation}} \\

Same hand ($27$)
& 1.218 / 1.410
& 1.389 / 1.467
& 1.340 / 1.558 \\

Different hands ($88$)
& \textbf{1.162 / 1.312}
& \textbf{1.121 / 1.283}
& \textbf{1.126 / 1.309} \\
\bottomrule
\end{tabular}
\end{table}

Uniform fusion already improves over every individual common-target estimate,
showing that collaboration becomes effective once observations from
heterogeneous wearable roles are mapped to the same physical target.
Fixed-weight fusion provides a further improvement by accounting for average
differences in observer reliability. Information fusion achieves the lowest median and mean ATE
for all three target devices, suggesting that event-dependent
weighting can better exploit variations in prediction
reliability across wearable observations.

The watch--phone relation provides an additional view of observation diversity.
Both subsets use the same information-fusion strategy. Different-hand
configurations achieve lower median and mean ATE for all three targets than
same-hand configurations. This observation is consistent with the hypothesis
that spatially more diverse wearable observations can provide more
complementary information about the same locomotion process.

\subsection{Discussion}

The results support the central premise of CoWear: effective collaboration
requires spatial correspondence before fusion. Heterogeneous wearable devices
observe the same locomotion process from different body locations, but their
self-trajectories are not directly interchangeable. Once observations are
mapped to a common physical target, however, multi-device fusion provides
consistent gains, with uncertainty-aware fusion further exploiting
event-dependent differences in observer reliability.

CoWear therefore complements conventional single-device inertial benchmarks by
enabling controlled study of cross-device correspondence and collaboration
through synchronized, device-aligned trajectories.

The current benchmark focuses on indoor walking with three consumer wearable
roles, and smartphone placement remains more constrained than in fully
unconstrained daily use. Future extensions should cover broader activities,
environments, body placements, and wearable modalities.

\section{Conclusion}

We presented \method, a device-aligned benchmark for collaborative inertial
localization using a smartphone, smartwatch, and smart glasses. By associating
each wearable IMU with its own motion-capture trajectory, CoWear preserves the
spatial correspondence between heterogeneous device observations and their
physical motion. Building on this correspondence, we formulate collaborative
localization as a common-target estimation problem, in which observations from
different body locations are first mapped to the motion of the same physical
target before fusion.

Experiments show that heterogeneous wearables can provide complementary
localization information despite following different local trajectories.
Common-target fusion consistently improves localization across different target
roles, while uncertainty-aware fusion and more diverse device placements
further improve collaborative performance. These results establish CoWear as a
benchmark for studying how spatially distinct consumer wearables can cooperate
for inertial localization and provide a foundation for future work on more
general collaborative wearable sensing.
\bibliographystyle{IEEEtran}
\bibliography{references}

\end{document}